\documentclass[sigconf]{acmart}

\AtBeginDocument{%
  }
\usepackage[linesnumbered,lined,boxed,commentsnumbered,ruled,vlined]{algorithm2e}

\copyrightyear{2023}
\acmYear{2023}
\setcopyright{acmlicensed}\acmConference[MM '23]{Proceedings of the 31st ACM International Conference on Multimedia}{October 29-November 3, 2023}{Ottawa, ON, Canada}
\acmBooktitle{Proceedings of the 31st ACM International Conference on Multimedia (MM '23), October 29-November 3, 2023, Ottawa, ON, Canada}
\acmPrice{15.00}
\acmDOI{10.1145/3581783.3612259}
\acmISBN{979-8-4007-0108-5/23/10}

\usepackage{balance}
\usepackage{multirow}
\usepackage{float}
\usepackage{adjustbox}
\acmSubmissionID{2436}

\begin{document}

\title{Semi-Supervised Panoptic
Narrative Grounding}

\author{Danni Yang}
\email{yangdanni@stu.xmu.edu.cn}
\affiliation{
  \institution{Key Laboratory of Multimedia Trusted Perception and Efficient Computing,\\
  Ministry of Education of China,\\
  Xiamen University,}
  \city{Xiamen}
  \state{Fujian}
  \country{China}
}

\author{Jiayi Ji}
\email{jjyxmu@gmail.com}
\authornote{Corresponding author.}
\affiliation{
  \institution{Key Laboratory of Multimedia Trusted Perception and Efficient Computing,\\
  Ministry of Education of China,\\
  Xiamen University,}
  \city{Xiamen}
  \state{Fujian}
  \country{China}
}

\author{Xiaoshuai Sun}
\email{xssun@xmu.edu.cn}
\affiliation{
  \institution{Key Laboratory of Multimedia Trusted Perception and Efficient Computing,\\
  Ministry of Education of China,\\
  Xiamen University,}
  \city{Xiamen}
  \state{Fujian}
  \country{China}
}

\author{Haowei Wang}
\email{wanghaowei@stu.xmu.edu.cn}
\affiliation{
  \institution{Key Laboratory of Multimedia Trusted Perception and Efficient Computing,\\
  Ministry of Education of China,\\
  Xiamen University,}
  \city{Xiamen}
  \state{Fujian}
  \country{China}
}

\author{Yinan Li}
\email{yinanlee@stu.xmu.edu.cn}
\affiliation{
  \institution{Key Laboratory of Multimedia Trusted Perception and Efficient Computing,\\
  Ministry of Education of China,\\
  Xiamen University,}
  \city{Xiamen}
  \state{Fujian}
  \country{China}
}

\author{Yiwei Ma}
\email{yiweima@stu.xmu.edu.cn}
\affiliation{
  \institution{Key Laboratory of Multimedia Trusted Perception and Efficient Computing,\\
  Ministry of Education of China,\\
  Xiamen University,}
  \city{Xiamen}
  \state{Fujian}
  \country{China}
}

\author{Rongrong Ji}
\email{rrji@xmu.edu.cn}
\affiliation{
  \institution{Key Laboratory of Multimedia Trusted Perception and Efficient Computing,\\
  Ministry of Education of China,\\
  Xiamen University,}
  \city{Xiamen}
  \state{Fujian}
  \country{China}
}

\renewcommand{\shortauthors}{Danni Yang et al.}

\begin{abstract}
Despite considerable progress, the advancement of Panoptic Narrative Grounding (PNG) remains hindered by costly annotations. In this paper, we introduce a novel Semi-Supervised Panoptic Narrative Grounding (SS-PNG) learning scheme, capitalizing on a smaller set of labeled image-text pairs and a larger set of unlabeled pairs to achieve competitive performance.
Unlike visual segmentation tasks, PNG involves one pixel belonging to multiple open-ended nouns. As a result, existing multi-class based semi-supervised segmentation frameworks cannot be directly applied to this task. To address this challenge, we first develop a novel SS-PNG Network (SS-PNG-NW) tailored to the SS-PNG setting. We thoroughly investigate strategies such as Burn-In and data augmentation to determine the optimal generic configuration for the SS-PNG-NW.
Additionally, to tackle the issue of imbalanced pseudo-label quality, we propose a Quality-Based Loss Adjustment (QLA) approach to adjust the semi-supervised objective, resulting in an enhanced SS-PNG-NW+. Employing our proposed QLA, we improve BCE Loss and Dice loss at pixel and mask levels, respectively. We conduct extensive experiments on PNG datasets, with our SS-PNG-NW+ demonstrating promising results comparable to fully-supervised models across all data ratios.
Remarkably, our SS-PNG-NW+ outperforms fully-supervised models with only 30\% and 50\% supervision data, exceeding their performance by 0.8\% and 1.1\% respectively. This highlights the effectiveness of our proposed SS-PNG-NW+ in overcoming the challenges posed by limited annotations and enhancing the applicability of PNG tasks. The source code is available at \url{https://github.com/nini0919/SSPNG}.
\end{abstract}




\maketitle

\section{Introduction}

The Panoptic Narrative Grounding (PNG) task is rapidly gaining prominence as a critical area of research in the multimodal domain~\cite{ma2022xclip,ma2023xmesh,wu-etal-2023-information,wu-etal-2023-cross2stra,fei-etal-2023-scene,wang2023nice}. This task aims to generate a pixel-level mask for each noun present in a given long sentence, providing a more fine-grained understanding compared to other cross-modal tasks, such as image captioning~\cite{xu2015show,vinyals2016show,cornia2020meshed,pan2020x,luo2021dual}, visual question answering~\cite{zhou2015simple,shih2016look,wu2017visual,kafle2017visual}, and referring expression comprehension/segmentation~\cite{luo2020cascade,cheng2021exploring,li2021bottom,liao2022progressive,liu2021cross,hui2020linguistic}. This level of detail sets it apart and opens up a wide range of potential applications, including fine-grained image editing~\cite{wang2021attribute,jiang2021talk} and fine-grained image-text retrieval~\cite{he2021cross,peng2022relation}.

\begin{figure}[t] 
\centering 
\includegraphics[width=1\columnwidth]{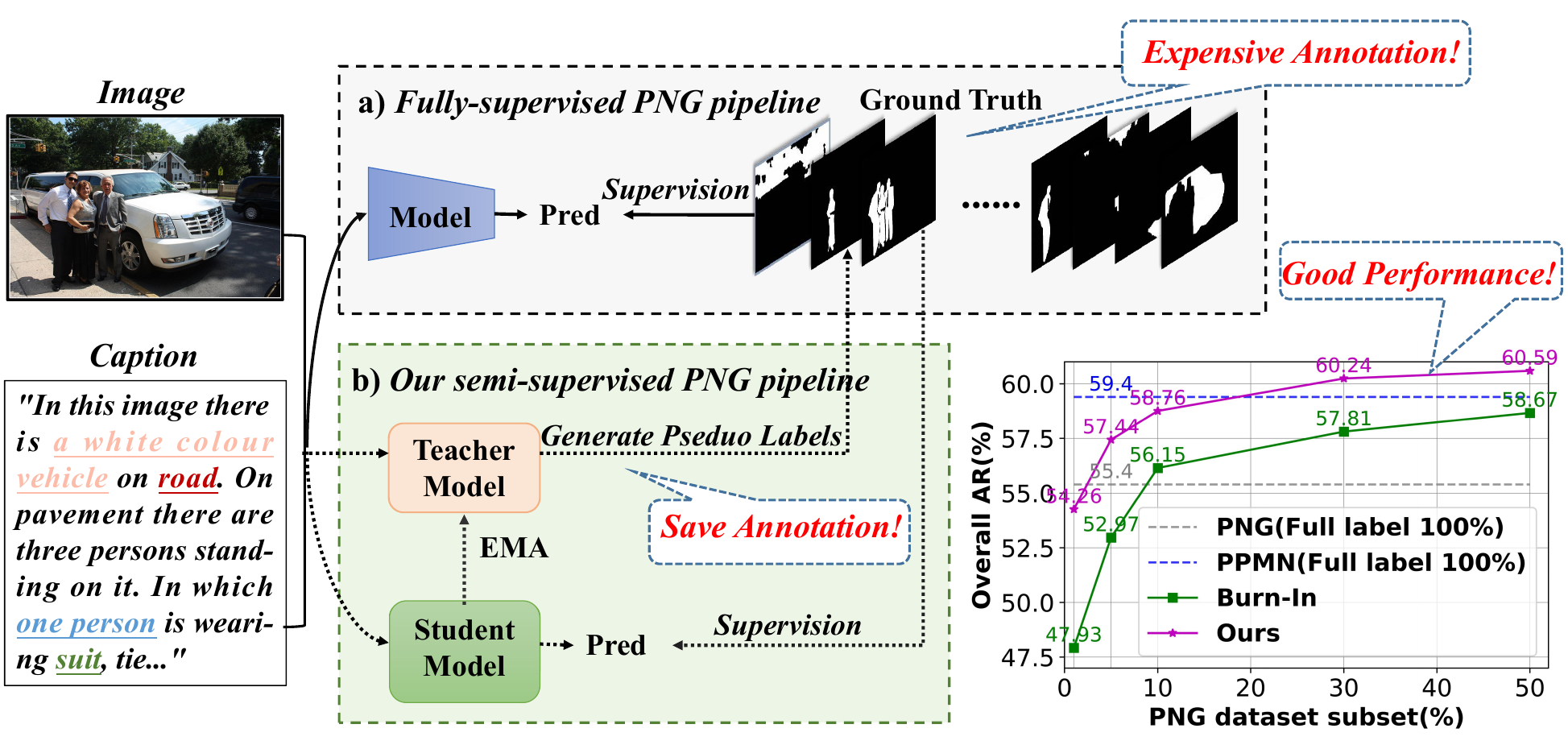}  
\caption{(a) A typical fully-supervised PNG pipeline. (b) Our proposed semi-supervised PNG pipeline.} 
\label{fig:intro}
\vspace{-10pt}
\end{figure}

Despite the recent significant advancements in the Panoptic Narrative Grounding (PNG) task, the need for detailed pixel-level annotations set it apart from other types of annotations, such as bounding boxes and categories. The precision required for PNG tasks entails substantial financial and human resource investments. Following the labeling budget calculation in \cite{kim2023devil}, on average, it takes approximately 79.1 seconds to segment a single mask. With each PNG example containing an average of 5.1 nouns requiring segmentation annotations~\cite{gonzalez2021panoptic}, this time expenditure increases to 403.4 seconds. This considerable constraint hampers dataset expansion and further limits model performance.
As a result, a natural inclination is to explore the potential of training models using a smaller number of image-text pairs with segmentation labels, in conjunction with a larger number of pairs without such labels, to achieve competitive performance. This approach is known as semi-supervised learning. Building upon this premise, we introduce a novel setting in this paper, termed Semi-Supervised Panoptic Narrative Grounding (SS-PNG)~\footnote{To maintain clarity and emphasize our methodological focus, we opt to use the term Semi-Supervised Panoptic Narrative Grounding, although it could be considered as Weakly Semi-Supervised in some aspects.}, as shown in Fig.~\ref{fig:intro}.

Semi-supervised learning has been employed in various vision tasks, such as semi-supervised object detection~\cite{sohn2020detection,liu2021unbiased,li2022pseco,wang2023consistent,zhou2021instant} and semi-supervised semantic segmentation~\cite{mittal2019semi,wang2022semi,st++,unimatch,zhen23augseg}, to alleviate the burden of manual annotation. However, applying these approaches directly to the SS-PNG task is challenging due to the unique characteristics of the task itself. While existing semi-supervised semantic segmentation methods primarily rely on pixel-level multi-classification and map them to corresponding categories, the PNG task presents a different scenario. In the context of PNG, there are two main differences. First, a single pixel can be associated with multiple nouns. Second, the categories of nouns are open-ended. These two fundamental distinctions render the traditional pixel-level multi-classification methods unsuitable for the PNG task, necessitating the development of novel techniques tailored to its specific requirements.

In light of these challenges, we propose a novel Semi-Supervised Panoptic Narrative Grounding Network (SS-PNG-NW) that effectively leverages unlabeled data through the use of pseudo-labels. Concretely, given an unlabeled image-text pair, we first obtain predictions from the model trained on labeled data and use the pixel-wise prediction as the ``ground truth'' to subsequently enhance the supervised model. Furthermore, we explore the effectiveness of conventional Burn-In strategies and various data augmentation techniques to identify the optimal configuration for the SS-PNG task.
We also recognize that the quality of pseudo-labels can vary, and high-quality pseudo-labels should have a more significant impact. To tackle this issue, we enhance the SS-PNG-NW to create the SS-PNG-NW+, utilizing a novel Quality-Based Loss Adjustment approach to refine the semi-supervised objective. Specifically, we first develop methods to assess the quality of pixel-level labels and mask-level labels. We then use the corresponding quality coefficients to adjust the Binary Cross-Entropy loss and Dice loss, respectively. Experimental results demonstrate that our proposed SS-PNG-NW+ can effectively exploit a limited amount of segmentation-labeled data to achieve competitive performance. As shown in Fig.~\ref{fig:intro}, with 1\%, 5\%, 10\%, 30\%, and 50\% of labeled data, our method achieves overall performance of 54.26\%, 57.44\%, 58.76\%, 60.24\%, and 60.59\%, respectively. Surprisingly, using only 30\% of labeled data, SS-PNG-NW+ surpasses the performance with 100\% of labeled data, highlighting the potential of our method in real-world applications.

In summary, our contributions are three-fold:
\begin{itemize}
\item We introduce a novel Semi-Supervised Panoptic Narrative Grounding (SS-PNG) setting, which reduces the dependency on annotated data, making it a more practical and cost-effective approach for the PNG task.
\item We propose an effective SS-PNG-NW that leverages pseudo-labels to utilize unlabeled data, and we explore various Burn-In strategies and data augmentation techniques to identify the optimal configuration for this task. Importantly, we investigate how to capitalize on high-quality pseudo-labels to further enhance the model's performance.
\item Extensive experiments demonstrate that our proposed SS-PNG-NW and SS-PNG-NW+ achieve competitive performance with a limited amount of segmentation-labeled data. Remarkably, our approach outperforms the performance achieved using 100\% labeled data when only 30\% of labeled data is used.%
\end{itemize}

\section{Related work}
\subsection{Panoptic Narrative Grounding}
The Panoptic Narrative Grounding (PNG) task aims to integrate natural language and visual information for more sophisticated scene perception. Specifically, the PNG task seeks to segment objects and regions in an image corresponding to nouns in its long text description. Numerous studies have been conducted on this task \cite{gonzalez2021panoptic, ding2022ppmn, wang2023towards}. González et al.~\cite{gonzalez2021panoptic} first introduced this new task, establishing a benchmark that includes new standard data and evaluation methods, and proposed a robust baseline method as the foundation for future work. To address the limitations of the previous two-stage approach, such as low-quality proposals and spatial details loss, Ding et al.~\cite{ding2022ppmn} proposed a one-stage Pixel-Phrase Matching Network that directly matches each phrase to its corresponding pixels and outputs panoptic segmentation. Concurrently, Wang et al.~\cite{wang2023towards} proposed a similar one-stage network for real-time PNG, but with a greater focus on the real-time performance of the model.

\subsection{Semi-Supervised Semantic Segmentation}
Manual pixel-level annotation for semantic segmentation is time-consuming and costly. Therefore, utilizing available unlabeled images to assist in learning segmentation models is of great value. Semi-supervised semantic segmentation tasks have recently developed rapidly, with many research works emerging~\cite{ouali2020semi,chen2021semi,zou2020pseudoseg,wang2022semi,yang2022revisiting}. Similar to the core concept of co-training~\cite{blum1998combining,zhou2005semi,zhang2014semi,peng2020deep}, CPS~\cite{chen2021semi} adopts dual models to supervise each other. As an extension of FixMatch~\cite{sohn2020fixmatch}, PseudoSeg~\cite{zou2020pseudoseg} extends weak consistency to strong consistency in segmentation scenarios and further applies a calibration module to refine the pseudo masks. U2PL \cite{wang2022semi} treats uncertain pixels as reliable negative samples to contrast against corresponding positive samples.
\begin{figure*}[]
  \centering
  \includegraphics[width=0.88\linewidth]{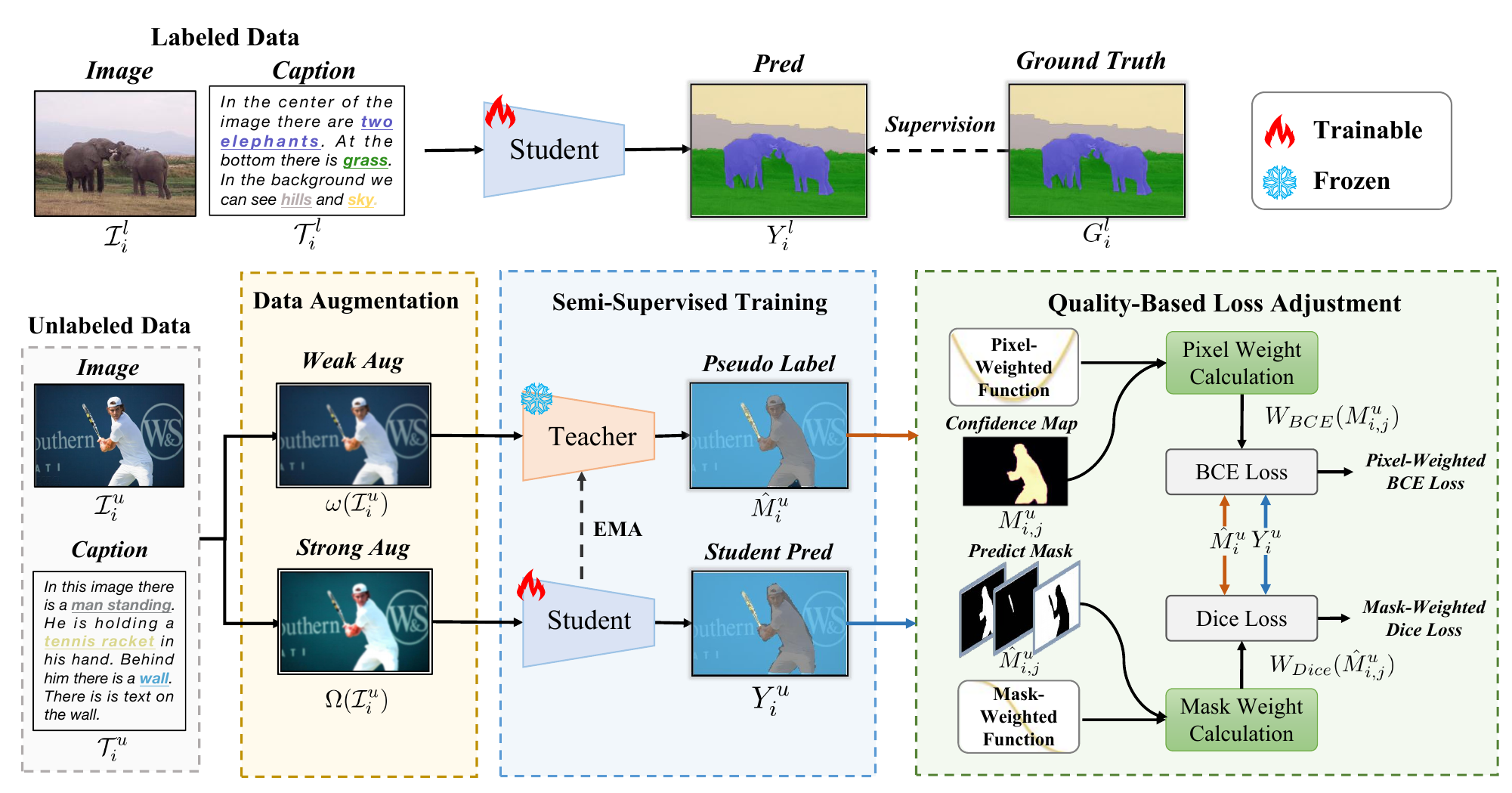}
  \caption{An overview of the SS-PNG-NW+. Our approach consists of a student network and a teacher network, with the latter being updated through Exponential Moving Average (EMA). 
  The proposed QLA approach is employed to prioritize high-quality pseudo-labels, facilitating pixel-level weighting for BCE loss and mask-level weighting for Dice loss.}
  \label{fig:framework}
  \vspace{-0.5em}
\end{figure*}

\section{SS-PNG Network (SS-PNG-NW)}
In this section, we first present the mathematical formulation of the SS-PNG task in Sec.~\ref{sec:overview}, as shown in Fig.~\ref{fig:framework}. Next, we explore the use of data augmentation strategies in Sec.~\ref{sec:data_aug}. Finally, we delve into our proposed two-stage semi-supervised training process designed for the SS-PNG task in Sec.~\ref{sec:sst}. 

\subsection{Task Definition}
\label{sec:overview}
In the Semi-Supervised Panoptic Narrative Grounding (SS-PNG) setting, we use a small labeled dataset, $\mathcal{D}_l=\left\{\left(\left(\mathcal{I}_i^l,  
\mathcal{T}_i^l\right),G_i^l\right)\right\}_{i=1}^{N^l}$ and a much
larger unlabeled set $\mathcal{D}_u=\left\{\left(\left(\mathcal{I}_i^u, \mathcal{T}_i^u\right),\varnothing
\right)\right\}_{i=1}^{N^u}$ 
, where $\mathcal{I}_i^l, \mathcal{I}_i^u$ is the $i$-th labeled image and $i$-th unlabeled image  , $\mathcal{T}_i^l, \mathcal{T}_i^u$ is the corresponding long narrative text, $G_i^l$ is the ground truth mask of $i$-th labeled image $\mathcal{I}_i^l$. ${N^l}$ and ${N^u}$ are the number
of labeled and unlabeled data, respectively, and commonly ${N^l} \ll {N^u}$. It is important to note that there are no ground truth mask labels in the unlabeled set $\mathcal{D}_u$, and the narrative texts are only used as inputs. Our final goal is to train a semi-supervised framework for the PNG that can effectively leverage a small portion of labeled data and a large portion of unlabeled data to achieve competitive performance.

\subsection{Data Augmentation Strategy}
\label{sec:data_aug}
Data augmentation plays a crucial role in enhancing the generalization and robustness of models in computer vision and natural language processing tasks, especially in semi-supervised learning~\cite{cubuk2019autoaugment,cubuk2019randaugment,hendrycks2019augmix,zoph2020learning,zhang2017mixup,zhong2020random,hendrycks2019augmix,devries2017improved}. SSL methods such as UDA~\cite{xie2020unsupervised} and FixMatch~\cite{sohn2020fixmatch} heavily rely on robust data augmentation techniques to utilize the abundant unlabeled data effectively. By ensuring consistent predictions under various input perturbations, data augmentation has emerged as a key driving factor in semi-supervised learning.
\begin{figure}[]
    \centering
    \includegraphics[width = 0.47\textwidth]{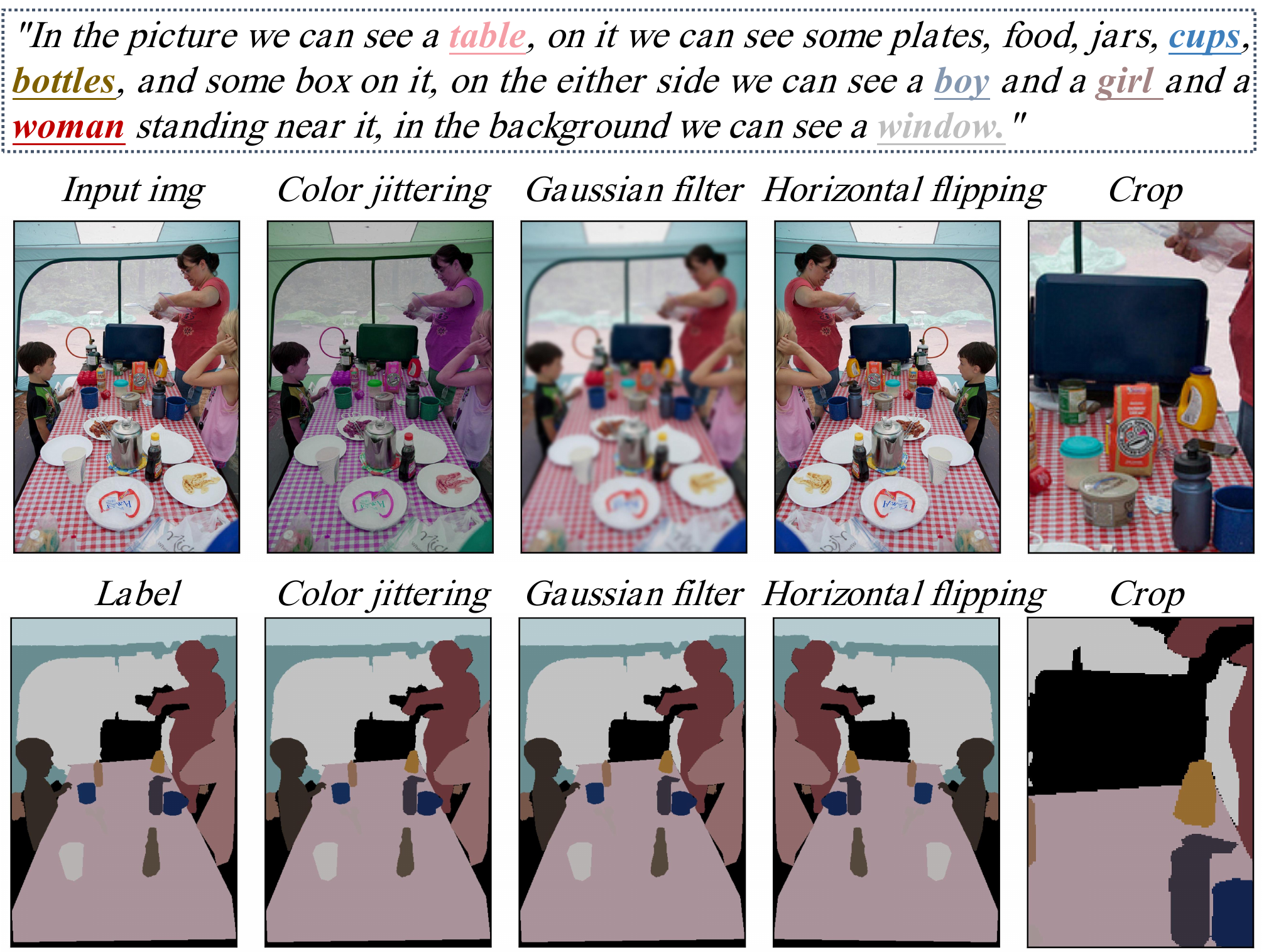}
    \caption{(a) First row: Input image and Augmented image. (b) Second row: Label and Augmented Label.}
    \label{fig:data_aug}
\end{figure}
In instance/semantic segmentation tasks~\cite{olsson2021classmix,yuan2021simple}, data augmentation techniques can be grouped into two types: i) methods requiring synchronous modification of the masks and original images, such as random flipping and random cropping; and ii) methods altering only the original images without affecting the labeled mask, including color jittering and Gaussian filter. Our primary focus is to investigate the effectiveness of these strategies for SS-PNG to develop a robust semi-supervised framework.

We design strong and weak augmentation approaches for semi-supervised PNG tasks, and search for the best augmentation schemes through experimentation. For unlabeled image input, our optimal scheme consists of Gaussian filter and horizontal flipping as weak augmentation, with color jittering added as strong augmentation. The teacher model receives weakly augmented unlabeled images, while the student model is fed with strongly augmented ones. 
It is worth noting that strong augmentation is built on top of weak augmentation. The weak and strong augmentations are represented by $\omega(\cdot)$ and $\Omega(\cdot)$, respectively. Fig.~\ref{fig:data_aug} demonstrates the data augmentation strategies employed in our experiments.
\subsection{Semi-Supervised Training for PNG}
\label{sec:sst}
\subsubsection{Burn-In Stage: A good initialization} 
\
\newline
A proper initialization is essential for both student and teacher models in SSL learning~\cite{liu2021unbiased,liu2022unbiased}, as the teacher model generates pseudo-labels to train the student model in later stages. To achieve this, we initially use PPMN~\cite{ding2022ppmn} as our Burn-In model for fully supervised training to obtain the prediction $Y_i^l \in \mathbb{R}^{N_i^l\times H_i^l\times W_i^l}$ of the $i$-th image $\mathcal{I}_i^l$:
\begin{equation}
Y_i^l=\mathcal{P}\left((Y_{i,1}^l, Y_{i,2}^l, \cdots, Y_{i,N_i^l}^l) \mid \Omega(\mathcal{I}_i^l),\left(\mathcal{T}_{i,1}^l,\mathcal{T}_{i,2}^l, \cdots, \mathcal{T}_{i,N_i^l}^l\right)\right),
\end{equation}
\noindent where $\mathcal{P}$ is the Burn-In model, $N_i^l$ represents the number of noun phrases in the i-th labeled  image $\mathcal{I}_i^l$. And $\mathcal{T}_{i,1}^l$ denotes the first noun phrase corresponding to the $i$-th labeled image $\mathcal{I}_i^l$, and $Y_{i,1}^l$is the prediction mask corresponding to the first noun phrase. $\Omega(\cdot)$ denotes the strong augmentation. $H_i^l$ and $W_i^l$ denote the height and width of the $i$-th labeled image.\par
Then we will use the ground truth $G_{i}^l \in \mathbb{R}^ {N_i^l \times H_i^l\times W_i^l}$ of the $i$-th image to supervise the prediction $Y_i^l$ with the loss $\mathcal{L}_{\text{sup}}$:
\begin{equation}
\mathcal{L}_{sup}(Y_i^l,G_i^l)=\frac{1}{N_i^l} \sum_{j=1}^{N_i^l} \mathcal{H}\left(G_{i,j}^l, {Y}_{i,j}^l\right),
\label{2}
\end{equation}
where $\mathcal{H}$ is loss function for the PNG task. Following the Burn-In stage, we replicate the well-trained model parameters to both teacher and student models in the mutual learning stage, preparing them for the subsequent training process. 

\subsubsection{Iterative Mutual Learning for Teacher-Student Convergence} 
\
\newline
\noindent \textbf{Step 1: Teacher Model Generates Pseudo-Labels} 
\
\newline
In our mutual learning process, we first feed weakly augmented unlabeled images and the corresponding descriptions into the teacher model to generate confidence maps $M_i^u \in \mathbb{R}^{N_i^u\times H_i^u\times W_i^u}$ of the $i$-th unlabeled image  $\mathcal{I}_i^u$, which are then applied to guide the student model's output in the subsequent step:
\begin{equation}
M_i^u=\mathcal{P}_{t}\left((M_{i,1}^u, M_{i,2}^u, \cdots, M_{i,N_i^u}^u) \mid \omega(\mathcal{I}_i^u),\left(\mathcal{T}_{i,1}^u,\mathcal{T}_{i,2}^u, \cdots, \mathcal{T}_{i,N_i^u}^u\right)\right),
\end{equation}

\noindent where $\mathcal{P}_t$ is the teacher model,  $N_i^u$ represents the number of noun phrases in the $i$-th unlabeled  image. $\mathcal{T}_{i,1}^u$ denotes the first noun phrase corresponding to the $i$-th unlabeled image $\mathcal{I}_i^u$, and $ M_{i,1}^u$ is the corresponding confidence maps that generated by teacher model, $\omega(\cdot)$ denotes weak augmentations. $H_i^u$ and $W_i^u$ denote the height and width of the $i$-th unlabeled image.\par\par
Then the teacher model's one-hot encoded output for the $k$-th pixel corresponding to the $j$-th noun phrase of the $i$-th unlabeled image is encoded as followed to obtain pseudo-labels $\hat M_{i}^u$:
\begin{equation}
{\hat M_{i,j,k}^u}= \begin{cases}0 ,& {M_{i,j,k}^u} \leq 0.5 \\ 1, & {M_{i,j,k}^u}>0.5\end{cases}.
\label{one-hot}
\end{equation}
\noindent \textbf{Step 2: Student Model Learning from Pseudo-Labels.}
\
\newline
In step 2, we apply strong augmentation to the $i$-th unlabeled image and feed it to the student model, obtaining the mask the predictions $Y_i^u \in \mathbb{R}^{N_i^u\times H_i^u\times W_i^u}$ of the $i$-th unlabeled image. 
\begin{equation}
Y_i^u=\mathcal{P}_{s}\left((Y_{i,1}^u, Y_{i,2}^u, \cdots, Y_{i,N_i^u}^u) \mid \Omega(\mathcal{I}_i^u),\left(\mathcal{T}_{i,1}^u,\mathcal{T}_{i,2}^u, \cdots, \mathcal{T}_{i,N_i^u}^u\right)\right),
\end{equation}

\noindent where $\mathcal{P}_s$ is the student model, $Y_{i,1}^u$ is the student model's prediction mask corresponding to the first noun phrase of the $i$-th unlabeled image, $\Omega(\cdot)$ denotes strong augmentations.\par
Then the teacher model's one-hot encoded output $\hat M_i^u$ supervises the student's predictions $Y_i^u$ using unsupervised loss $\mathcal{L}_{\text{unsup}}$:
\begin{equation}
\mathcal{L}_{unsup}(Y_i^u,\hat M_i^u)=\frac{1}{N} \sum_{j=1}^{N_i^u} \mathcal{H}\left(\hat M_{i,j}^u, {Y}_{i,j}^u\right),
\end{equation}

\noindent where $\mathcal{H}$ is loss function for the PNG task.

\noindent \textbf{Step 3: Stable Teacher Model Update with Exponential Moving Average (EMA).}
\
\newline
To ensure stable pseudo-labels, we avoid direct gradient-based updates to the teacher model's parameters. Instead, we use Exponential Moving Average (EMA) to create a more reliable model by calculating a weighted average of the previous model parameters and newly updated parameters. EMA has proven effective in many existing works~\cite{kingma2014adam,ioffe2015batch,he2020momentum,grill2020bootstrap,tarvainen2017mean}. By utilizing EMA, the teacher model's accuracy and stability are enhanced, making it more suitable for training and inference tasks. The formula is given as follows:
\begin{equation}
\theta_t \leftarrow \alpha \theta_t+(1-\alpha) \theta_s,
\label{ema}
\end{equation}
where $\theta_s $ is the parameters of the student model, $\theta_t$ denotes the parameters of the teacher model, and $\alpha$ is the decay coefficient of EMA, which is typically set to a small value, such as 0.99 in our experiments.

\section{SS-PNG NETWORK Plus (SS-PNG-NW+)}
\label{sec:soft_weight}
In the SS-PNG setting, the quality of pseudo-labels generated by the teacher model varies. When using these labels to guide the student model, it is essential to take quality information into account. In this section, we discuss how to leverage the quality information of pseudo-labels to improve model training. First, we introduce the general objective used in the SS-PNG in Sec.\ref{sec:loss}. Next, we explore how to utilize pixel-level and mask-level quality assessment methods to refine BCE loss and Dice loss, respectively, in Sec.\ref{sec:weitghted_loss}.

\subsection{Objective of SS-PNG Framework \label{sec:loss}}
\
\newline
In this paper, we approach panoptic narrative grounding as a segmentation task. Consequently, we adopt Binary Cross-Entropy (BCE) loss and Dice loss~\cite{milletari2016v} as our segmentation loss functions, following the precedent set by previous research~\cite{gonzalez2021panoptic,ding2022ppmn,wang2023towards}. BCE loss, a widely-used binary classification loss function, quantifies the distance between two probability distributions in binary classification problems. We define the BCE loss separately for the training of labeled and unlabeled data as $\mathcal{L}_{BCE}^l$ and $\mathcal{L}_{BCE}^u$, respectively, which are expressed as follows:
\begin{equation}
\mathcal{L}_{BCE}^l(Y_i^l,G_i^l)= -\frac{1}{N_i^l H_i^l W_i^l}  \sum_{j=1}^{N_i^l} \sum_{k=1}^{H_i^l \times W_i^l} L_{BCE}\left(G_{i,j,k}^l, {Y}_{i,j,k}^l\right),
\end{equation}

\begin{equation}
\mathcal{L}_{BCE}^u(Y_i^u,\hat M_i^u)=-\frac{1}{N_i^u H_i^u W_i^u}  \sum_{j=1}^{N_i^u} \sum_{k=1}^{H_i^u \times W_i^u} L_{BCE}\left(\hat  M_{i,j,k}^u, {Y}_{i,j,k}^u\right),
\label{eq:bceu}
\end{equation}
where $G_{i,j,k}^l, \hat M_{i,j,k}^u$ represent the ground truth/pseudo-label of the $k$-th pixel of the $j$-th noun phrase in the $i$-th labeled/unlabeled image, while $M_{i,j,k}^l, M_{i,j,k}^u$ are the model's predicted masks for the $i$-th labeled/unlabeled image. $L_{BCE}$ is the original BCE loss. \par
In contrast, Dice loss evaluates the similarity between predicted segmentation results and ground truth at the mask level. Similarly, we define the Dice loss separately for the training of labeled data and unlabeled data as $\mathcal{L}_{Dice}^l$ and $\mathcal{L}_{Dice}^u$, which are formulated as:\par
\begin{equation}
    \mathcal{L}_{Dice}^l(Y_i^l,G_i^l)=\sum_{j=1}^{N_i^l}\left(1-\frac{2|Y_{i,j}^l\bigcap G_{i,j}^l|}{|Y_{i,j}^l|+|G_{i,j}^l|}\right),
    \label{loss:dice}
\end{equation}
\begin{equation}
    \mathcal{L}_{Dice}^u(Y_i^u,\hat M_i^u)=\sum_{j=1}^{N_i^u}\left(1-\frac{2|Y_{i,j}^u\bigcap \hat M_{i,j}^u|}{|Y_{i,j}^u|+|\hat M_{i,j}^u|}\right),
    \label{eq:DiceU}
\end{equation}
Combining these two loss functions with different characteristics has been proven to improve model performance. Therefore, we obtain the supervised and unsupervised losses as follows:
\begin{equation}
\mathcal{L}_{sup}=\lambda_{1}^l\mathcal{L}_{BCE}^l+\lambda_{2}^l \mathcal{L}_{Dice}^l,
\end{equation}
\begin{equation}
\mathcal{L}_{unsup}=\lambda_{1}^u\mathcal{L}_{BCE}^u+\lambda_{2}^u \mathcal{L}_{Dice}^u,
\end{equation}
\noindent where $\lambda_{1}^l$ and $\lambda_{2}^l$ are the hyperparameters of supervised BCE loss and Dice loss, while $\lambda_{1}^u$ and $\lambda_{2}^u$ are the hyperparameters of unsupervised BCE loss and Dice loss, respectively. \par
In summary, our total training loss is expressed as follows:
\begin{equation}
\mathcal{L}=\mathcal{L}_{sup}+\lambda_{unsup} \mathcal{L}_{unsup},
\label{eq:unsupervised}
\end{equation}
\noindent where $\lambda_{unsup}$ is the hyperparameter of unsupervised loss $\mathcal{L}_{unsup}$.

\subsection{Quality-Based Loss Adjustment Approach (QLA)\label{sec:weitghted_loss}}
In summary, the BCE loss operates at the pixel level, while Dice loss focuses on the mask level. Therefore, in this section, we adopt two different pseudo-label quality assessment methods tailored to these two distinct loss functions.
\subsubsection{Pixel-wise Weight Adjustment for BCE Loss.}
\begin{figure}[]
  \centering
  \includegraphics[width=\linewidth]{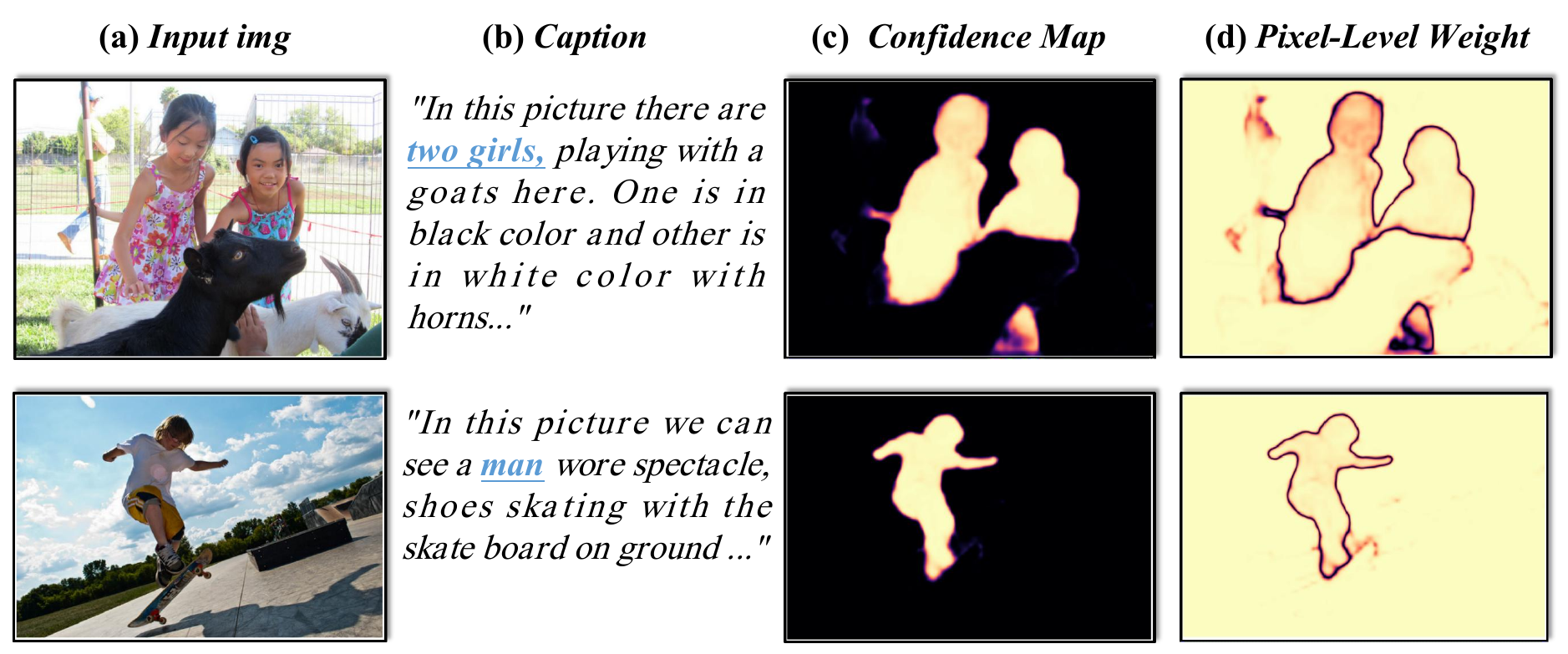}
  \caption{Examples of the proposed pixel-level weight are presented, which include (a) the original images, (b) the corresponding captions, (c) the confidence maps of the pseudo-labels, and (d) the pixel-level weights. Within the pixel weight, darker areas represent less reliable labels with poorer quality, predominantly found along the object edges. }
  \label{fig:pixelWeight}
  \vspace{-10pt}
\end{figure}

\
\newline
Since BCE loss is computed on a per-pixel basis, it is considered a pixel-level loss. Intuitively, when the output is close to 0 or 1, the pixel is more certain to be either background or foreground, resulting in a higher-quality label for that pixel. When the probability is closer to 0.5, the model's prediction for that pixel becomes more ambiguous. Therefore, when calculating BCE loss, we believe that not all pixels should be treated equally; instead, higher-quality labels should have higher weights. As such, we need to design an algorithm to assess the quality of each pixel. In this paper, we use a function to map the quality directly to a value between 0 and 1, reflecting the quality of each pixel:
\begin{equation}
W_{BCE}(M_{i,j,k}^u)=\beta-\frac{1}{\sqrt{2 \pi} \sigma} \exp \left(-\frac{(M_{i,j,k}^u-\mu)^2}{2 \sigma^2}\right),
\label{eq:wbce}
\end{equation}
where $\beta=1.3$, $\mu=0.5$, and $\sigma=0.1$. As shown in Eq.~\ref{eq:wbce}, the function approaches 0 when the probability is close to 0.5. The results are illustrated in Fig.~\ref{fig:pixelWeight}, revealing that low-quality pixels are primarily located at the edges. Lastly, the pixel-level BCE loss in Eq.~\ref{eq:bceu} can be rewritten as:
\begin{footnotesize}
\begin{equation}
    \mathcal{L}_{BCE}^u(Y_i^u,\hat M_i^u)=-\frac{1}{N_i^uH_i^uW_i^u}  \sum_{j=1}^{N_i^u} \sum_{k=1}^{H_i^u \times W_i^u} W_{BCE}(M_{i,j,k}^u) *L_{BCE}\left(\hat  M_{i,j,k}^u, {Y}_{i,j,k}^u\right).
    \label{loss:pixel-bce}
\end{equation} 
\end{footnotesize}

\subsubsection{Mask-wise Weight Adjustment for Dice Loss}
\begin{figure}[]
    \centering
    \includegraphics[width = 0.48\textwidth]{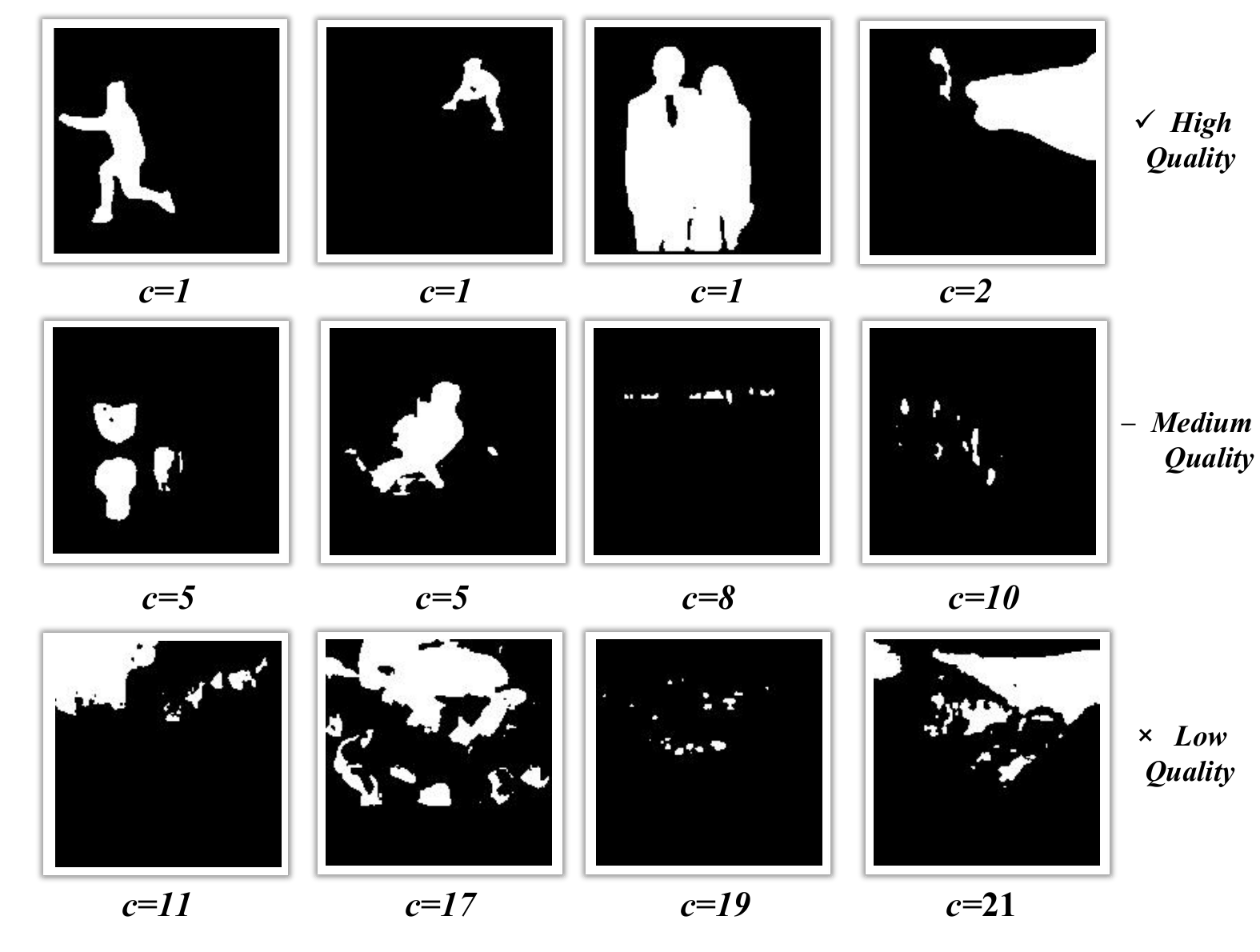}
    \caption{The visualization of mask connectivity and its relationship with the quality of pseudo mask labels. As the connectivity of the mask decreases, the quality of the corresponding pseudo-labels increases, where $\emph{c}$ represents the number of connected regions of this mask.}
    \label{fig:maskweight}
\end{figure}

\
\newline
Unlike BCE loss, Dice loss\cite{milletari2016v} considers the predicted and ground truth masks as a whole and emphasizes the overlapping areas between them. This characteristic of Dice loss makes it more suitable as a mask-level loss. Consequently, we have designed mask-level soft weights for the Dice loss.
As illustrated in Fig.~\ref{fig:maskweight}, the connectivity, or the number of connected components, is a good indicator of the quality of pseudo mask labels. The higher the connectivity, the lower the quality of the label. When the connectivity is 1, the label often has high quality.  We used the \emph{measure} function\footnote{\url{https://github.com/scikit-image/scikit-image/tree/main/skimage/measure.}} $C(\cdot)$ in the skimage library to compute the connectivity. Then we design a function to map the quality directly to a value between 0 and 1,
reflecting the quality of each mask:
\begin{equation}
W_{Dice}(\hat M_{i,j}^u)= \frac{1}{1+e^{C\left(\hat M_{i,j}^u\right)-\tau}},
\label{dd}
\end{equation}
where $\tau$ is the hyperparameter, used to adjust the translation magnitude of the curve. Considering that the connectivity of all pseudo-labels is relatively large at the beginning of training, the initial value of $\tau$ is also set to be large (20), and it decreases by 5 every 3k steps during training.
Finally, the Dice loss in Eq.~\ref{eq:DiceU} is reformulated as:
\begin{equation}
    \mathcal{L}_{Dice}^u(Y_i^u,\hat M_i^u)=\sum_{j=1}^{N_i^u} W_{Dice}(\hat M_{i,j}^u)*\left(1-\frac{2|Y_{i,j}^u\bigcap \hat M_{i,j}^u|}{|Y_{i,j}^u|+|\hat M_{i,j}^u|}\right),
    \label{loss:dice}
\end{equation}

\subsubsection{KL Divergence}
\
\newline
In addition, we also introduce KL divergence\cite{Kullback1951kl} for this task. The KL loss can be used to measure the difference between two probability distributions. When using pseudo-labels generated by the teacher model to supervise the student model, the probabilities are converted into 0-1 mask values through a threshold, as we do with BCE loss and Dice loss. This operation will result in the loss of a lot of information. Therefore, using the KL divergence method to directly extract the probability distribution from the teacher model is a perfect complement to the above two losses, as shown below:
\begin{equation}
\begin{aligned}
    \mathcal{L}_{KL}^u(Y_i^u, M_i^u)&=\mathcal{D}_{KL}( M_i^u,Y_i^u)\\
    &=\frac{1}{N_i^uH_i^uW_i^u}  \sum_{j=1}^{N_i^u} \sum_{k=1}^{H_i^u \times W_i^u} M_{i,j,k}^u \log \frac{ M_{i,j,k}^u}{{Y}_{i,j,k}^u},
\end{aligned}
\end{equation}

\noindent Finally, the unsupervised objective in Eq.~\ref{eq:unsupervised} is re-derived as follows:
\begin{equation}
\mathcal{L}_{unsup}=\lambda_{1}^u\mathcal{L}_{BCE}^u+\lambda_{2}^u \mathcal{L}_{Dice}^u+\lambda_{3}^u \mathcal{L}_{KL}^u,
\end{equation}
\noindent where $\lambda_{3}^u$ is the hyperparameters of KL loss.\par

\begin{table*}[htb]
\centering
\caption{Comparison of our proposed method and the state-of-the-art methods on the PNG benchmark. In our semi-supervised setting, \emph{F}1\% + \emph{U}99\% means using 1\% full-labeled data and 
 99\% unlabeled data. Other expressions follow in the same way.}
\vspace{-10pt}
\resizebox{0.78\textwidth}{!}{
    \begin{tabular}{l|cc|ccccc}
\toprule
\multirow{2}{*}{Method}
& \multirow{2}{*}{Label Types}
& \multirow{2}{*}{Budget(day)$\downarrow$}
& \multicolumn{5}{|c}{Segmentation Average Recall(\%) $\uparrow$} \\

&
&
& Overall
& Thing
& Stuff
& Single
& Plural  \\

\hline
\textit{Fully Supervised Models} & & & & & & & \\
\hline
PNG~\cite{gonzalez2021panoptic}
& \emph{F}100\% 
& 801.1 
& 55.4 
& 56.2 
& 54.3 
& 56.2 
& 48.8 \\

PPMN~\cite{ding2022ppmn}
& \emph{F}100\% 
& 801.1
& 59.4
& 57.2 
& 62.5 
& 60.0 
& 54.0 \\

MCN~\cite{luo2020multi}
& \emph{F}100\% 
& 801.1 
& 54.2
& 48.6
& 61.4
& 56.6
& 38.8 \\
EPNG~\cite{wang2023towards}
& \emph{F}100\% 
& 801.1 
& 49.7 
& 45.6
& 55.5
& 50.2
& 45.1 \\

\hline
\textit{Semi-Supervised Models}\\
\hline
SS-PNG-NW+ & \emph{F}1\% + \emph{U}99\% & 8.0 &54.26 &50.79&59.08	 &54.69	&  50.37 \\
SS-PNG-NW+ & \emph{F}5\% + \emph{U}95\% & 40.1 & 57.44& 54.18 & 61.98 &58.06 & 51.81 \\
SS-PNG-NW+ & \emph{F}10\% + \emph{U}90\% & 80.1 & 58.76 & 55.72 & 62.99 & 59.47 &52.33  \\
SS-PNG-NW+ & \emph{F}30\% + \emph{U}70\% & 240.3&60.24  &57.25  &64.40  &60.89  & 54.37 \\
SS-PNG-NW+ & \emph{F}50\% + \emph{U}50\% & 400.6 & \textbf{60.59} & \textbf{57.62}  & \textbf{64.71}  & \textbf{61.23} & \textbf{54.79} \\
\bottomrule
\end{tabular}
}
\label{tab:total_exp}
\end{table*}

\section{Experiment}
\subsection{Datasets and Evaluation}
\textbf{Datasets.} We verify the effectiveness of our proposed SS-PNG-NW and SS-PNG-NW+ on the Panoptic Narrative Grounding benchmark \cite{gonzalez2021panoptic}. We train our model on this PNG dataset and compare our model with existing fully supervised methods. The PNG dataset consists of image-text pairs, each containing an average of 5.1 objects per long narrative, including thing and stuff. And the objects include singular and plural, which makes visual-textual alignment more complex. 
\par 
\noindent \textbf{Evaluation.} 
We evaluate the performance of our model from two aspects: annotation budget and segmentation accuracy.
For the annotation budget, following the labeling budget calculation in \cite{kim2023devil}, it takes approximately 79.1 seconds to segment a single mask.  Therefore, we can estimate the annotation cost under different semi-supervised data ratios. For the segmentation accuracy, we adopt average recall to evaluate our SS-PNG-NW+. Specifically, we calculate the Intersection over Union (IoU)~\cite{jiang2018acquisition} between segmentation predictions and ground-truth masks for all evaluated noun phrases for different categories, including thing, stuff, singular and plural objects. And the performance is evaluated on the Teacher model.\par

\subsection{Implementation Details}
We adopt PPMN~\cite{ding2022ppmn} as the baseline of the panoptic narrative grounding network. To maintain consistency with PPMN, we adopt the same version of the backbone as PPMN for the feature extraction stage of both the visual modality and the linguistic modality. We implement our proposed SS-PNG-NW+ in PyTorch~\cite{paszke2019pytorch} and train it with batch size 12 for 12k iterations on 4 RTX3090 GPUs. Adam~\cite{kingma2014adam} is utilized as the optimizer. The learning rate is set to $1 \times 10^{-4}$. The loss hyperparameters $\lambda_{1}^l, \lambda_{2}^l, \lambda_{1}^u, \lambda_{2}^u, \lambda_{3}^u, \lambda_{unsup}$ are all set to 1.

\subsection{Comparison with State-of-the-Art Methods}
We compare our semi-supervised model with those fully-supervised models. The State-of-the-Art Methods include MCN~\cite{luo2020multi}, PNG~\cite{gonzalez2021panoptic}, EPNG~\cite{wang2023towards}, and PPMN~\cite{ding2022ppmn}. As shown in Tab.\ref{tab:total_exp}, compared to other fully-supervised methods (100\% labeled data), our method SS-PNG-NW+ with 30\% labeled data shows better performance than the existing SOTA method PPMN improved by +0.84\% (60.24\% vs. 59.4\%). Moreover, our method SS-PNG-NW+ with 50\% labeled data achieves the best performance which improved by +1.19\% (60.59\% vs. 59.4\%). Meanwhile, our semi-supervised framework saves annotation budget greatly. This means our method can achieve equally good results with fewer labeled data, greatly reducing the cost of annotation. 

\begin{table}[t]
\centering
\caption{Combinations of different data augmentations. W.A. and S.A. denote weak and strong augmentation, respectively. }
\vspace{-10pt}
\begin{tabular}{@{}cc|cccccc@{}}
\toprule
W.A. & S.A & Overall & Thing & Stuff & Single & Plural  \\ \midrule

-&-& 52.08 & 48.38 & 57.23& 52.51 & 48.21 \\
GF&CJ&53.15&49.61&58.07&53.58&49.24  \\
GF+HP&CJ &\textbf{53.22}& \textbf{49.68}&	\textbf{58.13}&\textbf{53.63}&\textbf{	49.58}		 \\
GF+HP+C&CJ &52.95 &49.46&57.80	&53.37	&49.12	\\
\bottomrule
\end{tabular}
\label{tab:DB1}
\end{table}


\begin{table}[]
\centering
\caption{Ablation study of soft weight adjustment
Approach on Pixel-wise weight and Mask-wise weight.}
\vspace{-10pt}
\begin{tabular}{@{}cc|ccccccc@{}}
\toprule
BCE & DICE  &Overall& Thing & Stuff & Single & Plural  \\ \midrule
-&-&53.22& 49.68&	58.13&53.63&	49.58\\ 
pixel&-&53.99&50.47&\textbf{58.90}&54.44&50.04\\
pixel&pixel&18.47 &11.30  & 28.45 &20.08&3.99 \\
-&mask&53.87&50.30&58.83&54.27&50.23\\
pixel&mask&\textbf{54.10}&\textbf{50.78}&58.78&\textbf{54.51}&\textbf{50.36 }\\ \bottomrule
\end{tabular}
\label{tab:weight}

\end{table}


\subsection{Ablation Study}
For a fair comparison, we conduct the ablation study of our method on the ``$\emph{F}1\% + \emph{U}99\%$'' setting, which refers to using 1\% labeled data and the remaining 99\% unlabeled data.
\subsubsection{Effectiveness of Data Augmentation .}
\
\newline
In Tab.~\ref{tab:DB1}, we attempt to explore the best strategy for our task using variants of data augmentation techniques. We consider different data augmentations for the segmentation task, including Gaussian filter (GF), horizontal flipping (HF), color jittering (CJ), cropping (C). The above data augmentations are randomly applied with a probability of 0.5. As shown 3-rd row, when we choose GF and HF for weak augmentation and CJ for strong augmentation, we achieve the best performance. Compared to no data augmentation, the performance is improved by +1.14\% (53.22\% vs. 52.08\%).


\begin{table}[]
\centering
\caption{Ablation study of KL loss.}
\vspace{-10pt}
\begin{tabular}{@{}ccc|cccccc@{}}
\toprule
BCE & DICE &KL & Overall & Thing & Stuff & Single & Plural  \\ \midrule
\checkmark&\checkmark&&54.10&50.78&58.78&54.51&50.36\\
\checkmark&\checkmark&\checkmark&\textbf{54.26} &\textbf{50.79}&\textbf{59.08}	 &\textbf{54.69}	&  \textbf{50.37}\\
\bottomrule
\end{tabular}
\label{tab:loss}
\end{table}


\begin{table}[]
\centering
\caption{Ablation study on different components.}
\vspace{-10pt}
\begin{tabular}{@{}ccc|ccccccc@{}}
\toprule
 
DA & SST& QLA & Overall & Thing & Stuff & Single & Plural  \\ \midrule
 &&& 46.57 & 42.64 & 52.03 & 47.46&38.49\\
\checkmark&&& 47.93 & 43.88 & 53.55 & 49.08 & 37.49\\
 &\checkmark &&52.08 & 48.38 & 57.23& 52.51 & 48.21 \\
\checkmark&\checkmark&&53.22& 49.68&	58.13&53.63&	49.58\\
\checkmark& \checkmark &\checkmark  & \textbf{54.26 }& \textbf{50.79} & \textbf{59.08} & \textbf{54.69} & \textbf{50.37}\\ \bottomrule
\end{tabular}
\label{tab:ablation}
\end{table}

\begin{figure}[t]
  \centering
  \includegraphics[width=0.88\linewidth]{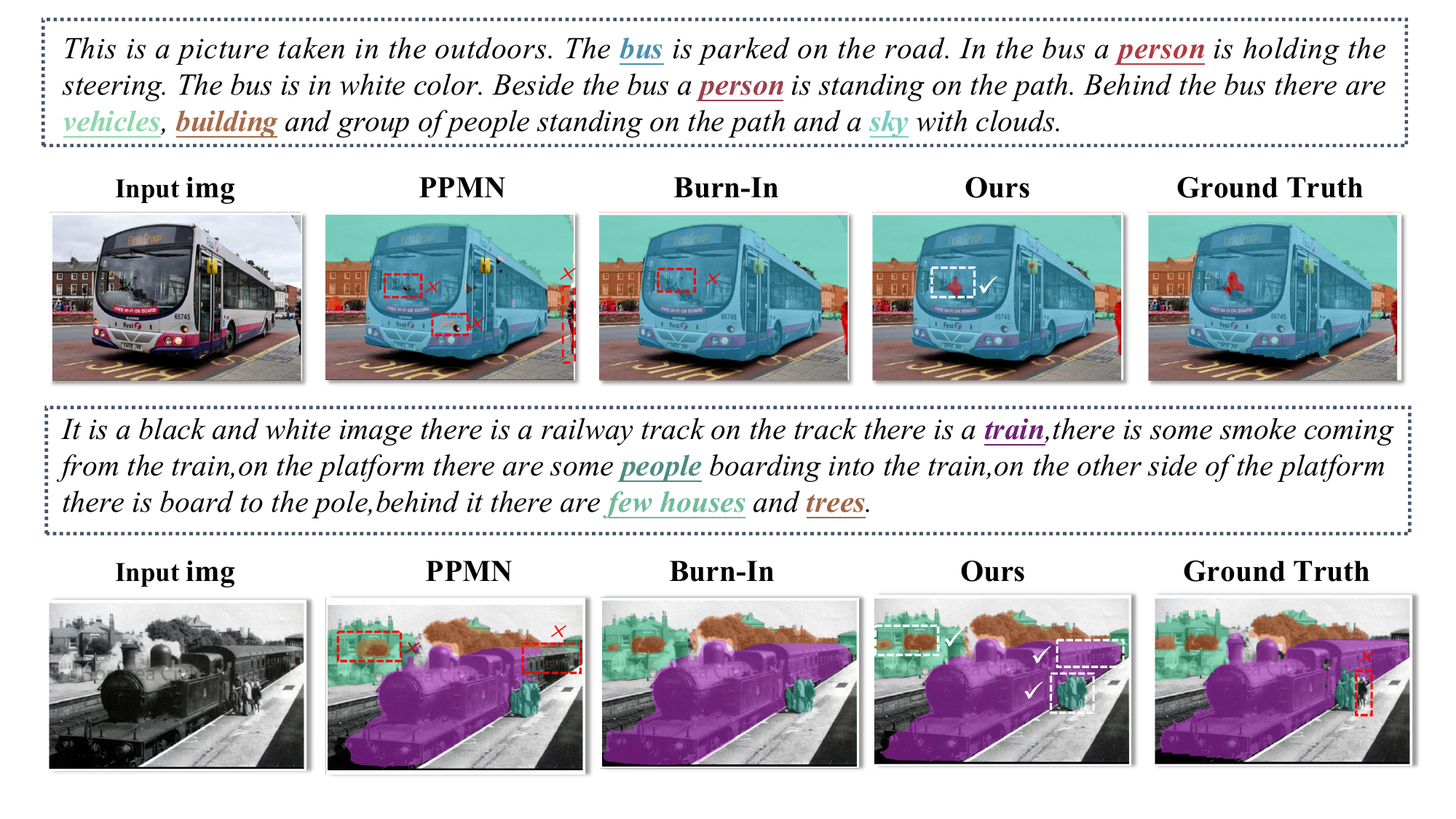}
  \vspace{-10pt}
  \caption{Qualitative analysis for our proposed SS-PNG-NW+. White dashed bounding boxes indicate accurate predictions of our model, while red dashed bounding boxes encompass the areas with inaccurate predictions of others. Here we use Burn-In and our semi-supervised model trained on 50\% labeled data for visualization.}
  \label{fig:visualization}
  \vspace{-10pt}
\end{figure}
\subsubsection{Effectiveness of Quality-Based Loss Adjustment Approach.}
\
\newline
In Tab.~\ref{tab:weight}, we first verify the effectiveness of our proposed Quality-Based loss adjustment approach, i.e., pixel-wise weight for BCE loss and mask-wise weight for Dice loss. The first row refers to the best result for the ablation of data augmentation. As shown in the 2-nd row, our designed pixel-wise weight for BCE loss is effective which improved by +0.77\% (53.99\% vs. 53.22\%).
And the results in the 3-rd row indicate that our pixel-level weight adjustment strategy on Dice loss did not yield satisfactory results (18.47\% vs. 53.99\%). Upon analysis, we find if we use the mapping function from confidence to Gaussian distribution, it will change the model prediction distribution. Therefore, as shown in the 4-th row, we try to add the mask-level soft weight on the Dice loss and find that it improved performance by +0.65\% (53.87\% vs. 53.22\%). And in the last row, we combine the two soft weight adjust approaches which can improve the overall AR by +0.88\% (54.10\% vs. 53.22\%).

In Tab.~\ref{tab:loss},  we further validate the effectiveness of the additional KL loss. It can be seen that the KL loss based on soft labels can slightly improve performance by +0.16\% (54.26\% vs. 54.10\%).

\subsubsection{Effectiveness of Different Components.}
\
\newline
We conduct experiments in Tab.~\ref{tab:ablation} to ablate each component of our framework step by step. The 1-st row indicates that only 1\% labeled data is used for supervised training, without adding any modules. And the 2-nd row means applying the strong data augmentation strategy (DA) on top of the labeled data used in the first line. As
shown in the 1-st and 2-nd rows, we find that data augmentation is effective for labeled data. The next step is to explore the effectiveness of our semi-supervised training (SST) for PNG. As is shown in the 3-rd row, compared to the 1-st row, the model performance has improved by +5.51\% (52.08\% vs. 46.57\%) due to the emergence of our semi-supervised training (SST). Then on the basis of the semi-supervised framework, we further enhance our designed data augmentation strategies and find that the performance is further improved by +1.14\% (53.22\% vs. 52.08\%), which further verified the effectiveness of data augmentation (DA) for the semi-supervised PNG task. In the final step, we incorporate our Quality-Based loss adjustment approach (QLA) and KL loss. We find that the model performance improved by +1.04\% (54.26\% vs. 53.22\%).

\subsection{Qualitative Analysis}
As shown in Fig.~\ref{fig:visualization}, we present some typical grounding results of our SS-PNG-NW+ compared to the Burn-In model, the PPMN, and the ground truth. Surprisingly, our proposed semi-supervised framework effectively corrects errors made by the PPMN and Burn-In model. In the first example, our model correctly identified the bus driver as the ``person holding the steering'' while PPMN and Burn-In model failed to do so. In the second example, the PPMN model did not perform well in identifying the back of the train and the tree, while our model correctly predicted them. Moreover, our model corrected the errors of ground truth for not fully recognizing ``the people on the platform''.

\section{Conclusion}
In this work, we present a novel Semi-Supervised Panoptic Narrative Grounding (SS-PNG) learning scheme to tackle the challenges posed by the expensive annotation process in PNG. We initially establish a dedicated SS-PNG Network (SS-PNG-NW) designed specifically for the SS-PNG setting. We proceed to comprehensively examine strategies such as Burn-In and data augmentation to identify the most suitable generic configuration for the SS-PNG-NW. Furthermore, we introduce the Quality-Based Loss Adjustment (QLA) approach to refine the semi-supervised objective, promoting more significant attention to high-quality pseudo-labels, thereby creating an improved SS-PNG-NW+. Our extensive experimental evaluations demonstrate that the proposed SS-PNG-NW+ achieves performance on par with fully supervised models while significantly reducing annotation expenses.

\begin{acks}
This work was supported by National Key R\&D Program of China (No.2022ZD0118201), the National Science Fund for Distinguished Young Scholars (No.62025603), the National Natural Science Foundation of China (No. U21B2037, No. U22B2051, No. 62176222, No. 62176223, No. 62176226, No. 62072386, No. 62072387, No. 62072389, No. 62002305 and No. 62272401), China Postdoctoral Science Foundation (No.2023M732948), and the Natural Science Foundation of Fujian Province of China (No.2021J01002,  No.2022J06001).
\end{acks}

\bibliographystyle{ACM-Reference-Format}
\balance
\bibliography{acmart}

\clearpage

\appendix
\section{Additional Related work}
\noindent\textbf{Semi-supervised Learning.}
Semi-supervised learning is a machine learning approach that utilizes a small amount of labeled data and a large amount of unlabeled data for training. It has been widely applied in computer vision and natural language processing. Semi-supervised learning has two typical paradigms: consistency regularization~\cite{bachman2014learning, french2019semi, ouali2020semi,xu2021dash} and entropy minimization~\cite{grandvalet2004semi,saito2019semi, wu2021semi, chen2021semisupervised}. And data augmentation plays an important role in semi-supervised learning, as it improves model generalization and robustness when applied to unlabeled data. Several works have presented augmentation techniques for semi-supervised learning, including CutOut~\cite{devries2017improved}, CutMix~\cite{yun2019cutmix}, and ClassMix~\cite{olsson2021classmix}.

\section{Labeling Budget Calculation}
The PNG dataset encompasses 133,103 training images and 8,380 testing images, with respective mask annotations numbering 875,073 and 56,531. In accordance with the labeling budget calculation detailed in \cite{kim2023devil}, the average time required to segment a single mask is approximately 79.1 seconds. Given this information, we compute the annotation budgets for training datasets under our semi-supervised configuration and the fully supervised scenario as follows:
\begin{itemize}
\item $\emph{F} 1\% + \emph{U} 99\%$:  875,073  × 0.01 × 79.1 / 60 / 60 / 24 = 8.0 day
\item $\emph{F} 5\% + \emph{U} 95\%$: 875,073  × 0.05 × 79.1 / 60 / 60 / 24 = 40.1 day
\item $\emph{F} 10\% + \emph{U} 90\%$: 875,073  × 0.1 × 79.1 / 60 / 60 / 24 = 80.1 day
\item $\emph{F}30\% + \emph{U}70\%$: 875,073  × 0.3 × 79.1 / 60 / 60 / 24 = 240.3 day
\item $\emph{F}50\% + \emph{U}50\%$: 875,073  × 0.5 × 79.1 / 60 / 60 / 24 = 400.6 day
\item $\emph{F}100\%$: 875,073 × 79.1 / 60 / 60 / 24 = 801.1 day
\end{itemize}

\noindent These underscore the significant reduction in annotation costs achieved by our proposed semi-supervised framework. This cost-efficiency is particularly impactful when considering the transfer of this framework to specific domains where labeling is prohibitively expensive.

\section{Entropy-Based Dropout Strategy \emph{v.s.} Quality-Based Loss Adjustment}
To further investigate how to leverage the confidence information of pseudo-labels, we explore not only the Quality-Based Loss Adjustment (QLA) approach, but also an additional filtering strategy known as the Entropy-Based Dropout Strategy (EDS). The central idea of EDS is to avoid misleading the model during training by filtering out pseudo-labels with low quality. In our study of EDS, we filter out 20\% of labels with high entropy, and document the results in Tab.~\ref{tab:twokind}. As can be seen, both strategies result in performance enhancements. Compared to EDS, QLA exhibits superior performance (54.26 vs. 53.75) because it utilizes more information; not only are high-quality labels exploited, but low-quality labels are also put to use.

\section{Effectiveness of SS-PNG-NW and SS-PNG-NW+}
In Tab.~\ref{tab:size}, we present the performance comparison between the Burn-In model and our semi-supervised model under different labeled data configurations. Burn-In model means only using labeled data for supervised training. Our model achieves improvements of 6.33\%, 4.47\%, 2.61\%, 2.43\% and 1.92\%, with 1\%, 5\%, 10\%, 30\% and 50\% labeled data, respectively. It can also be observed that the less labeled data available, the more significant the performance gain of our model.

Furthermore, we undertake comparative experiments involving SS-PNG-NW and SS-PNG-NW+ in Tab.~\ref{tab:size}. It is evident that both SS-PNG-NW and SS-PNG-NW+ achieve competitive results when compared with fully supervised models, which reinforces the validity of our semi-supervised framework. Additionally, it can be discerned that SS-PNG-NW+ consistently surpasses SS-PNG-NW across all configurations, substantiating the effectiveness of our proposed Quality-Based Loss Adjustment (QLA) module.

\begin{table}[]
\centering
\caption{Ablation of Entropy-Based Dropout Strategy (EDS) or Quality-Based Loss Adjustment (QLA).}
\vspace{-10pt}
\begin{tabular}{@{}c|cccccc@{}}
\toprule
Selection Type & Overall & Thing & Stuff & Single & Plural  \\ \midrule
-&53.22& 49.68 &58.13 &53.63& 49.58\\
EDS&53.75&50.21&58.67&54.20&49.69\\
QLA&\textbf{54.26}&\textbf{ 50.79}&\textbf{ 59.08}& \textbf{54.69}& \textbf{50.37}\\
\bottomrule
\end{tabular}
\label{tab:twokind}
\end{table}

\begin{table}[]
\centering
\caption{Comparison between SS-PNG-NW vs. SS-PNG-NW+ under different labeled sizes.}
\vspace{-10pt}
\begin{adjustbox}{width=0.48\textwidth}
\begin{tabular}{@{}cc|ccccccc@{}}
\toprule
Size & Model &Overall& Thing & Stuff & Single & Plural  \\ \midrule
1\%&Burn-In&47.93 & 43.88 & 53.55 & 49.08 & 37.49 \\
1\%&SS-PNG-NW&53.22& 49.68 &58.13& 53.63& 49.58 \\
1\%&SS-PNG-NW+&\textbf{54.26} &\textbf{50.79}&\textbf{59.08}	 &\textbf{54.69}	&  \textbf{50.37}&\\
\bottomrule
5\%&Burn-In&52.97& 49.22&	58.18&54.03&	43.46 \\
5\%&SS-PNG-NW& 56.94&53.50  &61.72 &57.51 &51.75	  \\
5\%&SS-PNG-NW+&\textbf{57.44}& \textbf{54.18} & \textbf{61.98} &\textbf{58.06} & \textbf{51.81} &\\
\bottomrule
10\%&Burn-In&56.15& 52.91&	60.66&56.93&	49.12 \\
10\%&SS-PNG-NW&58.50 &55.30  &62.94 &59.09 & \textbf{53.13} \\
10\%&SS-PNG-NW+&\textbf{58.76} & \textbf{55.72} & \textbf{62.99} & \textbf{59.47} & 52.33 & \\
\bottomrule
30\%&Burn-In&57.81& 54.20&	62.82 &58.72&	49.54 \\
30\%&SS-PNG-NW&59.39&56.21  &63.79  &59.98 &	53.98  \\
30\%&SS-PNG-NW+&\textbf{60.24 } &\textbf{57.25}  &\textbf{64.40 } &\textbf{60.89}  & \textbf{54.37}  \\
\bottomrule
50\%&Burn-In&58.67& 55.84&	62.59 & 59.25&	53.40 \\
50\%&SS-PNG-NW&59.98 &56.88  &64.30  &60.59 & 54.47\\
50\%&SS-PNG-NW+&\textbf{60.59} & \textbf{57.62}  & \textbf{64.71}  & \textbf{61.23} & \textbf{54.79} & \\
\bottomrule
100\%&PPMN&59.4 & 57.2 & 62.5 & 60.0 & 54.0 &	 \\
\bottomrule
\end{tabular}
\end{adjustbox}
\label{tab:size}
\end{table}
\end{document}